\documentclass[letterpaper, 10 pt, conference]{ieeeconf}  

\IEEEoverridecommandlockouts                              

\usepackage{float}
\usepackage{mathtools}
\usepackage{amsmath} 
\usepackage{amssymb}  
\usepackage{esvect}
\usepackage{xcolor}
\usepackage{algorithm}
\usepackage{algpseudocode}
\usepackage{tikz}

\usepackage{amsthm}
\usepackage{cleveref}
\let\labelindent\relax
\usepackage{enumitem}
\setlist[enumerate]{label=(\arabic*)}
\newcounter{counter}

\newtheorem{proposition}[counter]{Proposition}

\crefname{equation}{Eq.}{Eqs.}

\def\@firstoftwo@second#1#2{%
	\def\temp##1.##2\@nil{##2}%
	\temp#1\@nil}
\newcommand\sref[1]{%
	\expandafter\@setref\csname r@#1\endcsname\@firstoftwo@second{#1}%
}

\title{\LARGE \bf
Rejecting Outliers in 2D-3D Point Correspondences from\\2D Forward-Looking Sonar Observations
}

\author{Jiayi Su$^{1}$, Shaofeng Zou$^{2}$, Jingyu Qian$^{1}$, Yan Wei$^{2}$, Fengzhong Qu$^{2,3}$ and Liuqing Yang$^{1,4}$
\thanks{*This work is in part supported by Natural Science Foundation of China Project \#U23A20339, Guangzhou Municipal Science and Technology Project \#2023A03J0011, and Guangdong Provincial Project \#2023ZDZX1037 and \#2023ZT10X009.}
\thanks{$^{1}$The Intelligent Transportation Thrust, The Hong Kong University of Science and Technology (Guangzhou), Guangzhou 511458, China. Emails: {\{sujyzju, qianjingyu\}@zju.edu.cn, lqyang@ust.hk}}
\thanks{$^{2}$The School of Integrated Circuits, Tsinghua University, Beijing 100084, China. Email: {zousf19@mails.tsinghua.edu.cn}}
\thanks{$^{3}$The Engineering Research Center of Oceanic Sensing Technology and Equipment, Ministry of Education, Ocean College, Zhejiang University, Zhoushan 316021, China, and the Provincial Key Laboratory of Cutting-edge Scientific Instruments R\&D and Application, Ocean College, Zhejiang University, Zhoushan 316021, China. Emails: {\{redwine447, jimqufz\}@zju.edu.cn}}
\thanks{$^{4}$The Hainan Institute of Zhejiang University, Sanya 572025, China.}
\thanks{$^{5}$The Internet of Things Thrust, The Hong Kong University of Science and Technology (Guangzhou), Guangzhou 511458, China, and the Department of Electronic and Computer Engineering, The Hong Kong University of Science and Technology, Hong Kong SAR 999077, China.}
}

\usepackage{eso-pic}

\AddToShipoutPictureBG*{%
	\AtPageUpperLeft{%
		\put(60,-30){%
			\parbox{0.8\paperwidth}{\centering
				\footnotesize
				This paper has been accepted for presentation at the IEEE/RSJ IROS 2025 Hangzhou conference.
				Please cite the paper as:\\J. Su, et al. "Rejecting Outliers in 2D-3D Point Correspondences from 2D Forward-Looking Sonar Observations."\\2025 IEEE/RSJ International Conference on Intelligent Robots and Systems (IROS). IEEE, 2025, pp. 1-8.\\
			}
		}
	}
}

\begin{document}

\maketitle
\thispagestyle{empty}
\pagestyle{empty}


\begin{abstract}
Rejecting outliers before applying classical robust methods is a common approach to increase the success rate of estimation, particularly when the outlier ratio is extremely high (e.g. 90\%). However, this method often relies on sensor- or task-specific characteristics, which may not be easily transferable across different scenarios. In this paper, we focus on the problem of rejecting 2D-3D point correspondence outliers from 2D forward-looking sonar (2D FLS) observations, which is one of the most popular perception device in the underwater field but has a significantly different imaging mechanism compared to widely used perspective cameras and LiDAR. We fully leverage the narrow field of view in the elevation of 2D FLS and develop two compatibility tests for different 3D point configurations: (1) In general cases, we design a pairwise length in-range test to filter out overly long or short edges formed from point sets; (2) In coplanar cases, we design a coplanarity test to check if any four correspondences are compatible under a coplanar setting. Both tests are integrated into outlier rejection pipelines, where they are followed by maximum clique searching to identify the largest consistent measurement set as inliers. Extensive simulations demonstrate that the proposed methods for general and coplanar cases perform effectively under outlier ratios of 80\% and 90\%, respectively.
\end{abstract}

\section{Introduction}
Correct correspondences are crucial for solving various computer vision problems, such as two-view geometry, camera resectioning, and simultaneous localization and mapping (SLAM) \cite{hartley2003multiple,tang2019gcnv2,mur2015orb}. However, in the real world, incorrect correspondences are inevitable due to various factors, including noisy measurements and the limited performance of matching algorithms \cite{szeliski2022computer}. As a result, robust estimation is essential for handling outliers \cite{bosse2016robust}. Popular robust estimation methods, including RANSAC \cite{hartley2003multiple} and graduated non-convexity (GNC) \cite{yang2020graduated}, can manage moderate outlier ratios, but they often fail when the outlier ratio becomes excessively high (e.g., 90\%) \cite{shi2021robin}. To address this problem, recent studies have adopted the paradigm of consensus maximization \cite{chin2017maximum} to achieve robust estimation under extremely high outlier ratios (e.g., $\ge$ 95\%) \cite{shi2021robin,forsgren2024group}. They designed a series of compatibility tests and constructed a compatibility graph based on the results. Then, the maximum clique was identified and treated as the inlier set. However, this approach requires in-depth analysis of task-specific characteristics \cite{yang2020teaser}. In the domain of 2D forward-looking sonar (2D FLS), limited research attention has been devoted to this area despite its significant demand.

2D FLS has gradually gained attention for its use in underwater surveys, benefiting from its photo-realistic imaging quality. Extensive research has been conducted to enable advanced sonar vision, aiming to facilitate robot autonomy. The core steps of sonar vision involve establishing correspondences; however, researchers often assume these correspondences are known \textit{a priori} or rely on simple methods that are effective only in controlled environments.
As a result, the community lacks in-depth discussions on rejecting outliers in correspondences. In this paper, we attempt to take the first step by introducing an outlier rejection method specifically designed for 2D FLS. 2D-3D point(-to-point) correspondences are preliminarily considered. The contributions of the paper are summarized as follows:
\begin{enumerate}
	\item We propose two compatibility tests specifically designed for 2D-3D point correspondences from 2D FLS observations. The first test, termed the pairwise length in-range test, is applicable in general cases. The second test, referred to as the coplanarity test, is tailored for coplanar 3D point configurations.
	\item Both tests are integrated into an outlier rejection pipeline similar to those described in \cite{shi2021robin,forsgren2024group} for evaluation. The simulation results demonstrate that, in general cases, the proposed method is robust to an 80\% outlier ratio, while in coplanar cases, it is robust to a 90\% outlier ratio.
\end{enumerate}

\section{Related Works}
\label{sec:related works}

\textbf{Robust estimation} methods for managing outliers can be broadly categorized into two main approaches. The first is M-estimation, which uses non-minimal solvers with robust cost functions to reduce outlier impact \cite{mactavish2015all,black1996unification,yang2020graduated}. However, it requires a good initial guess and struggles with outlier ratios above 80\%. The second approach employs consensus maximization \cite{chin2017maximum}, aiming to identify the most consistent inlier set. This includes fast heuristics like RANSAC \cite{hartley2003multiple} and its variants \cite{chum2005matching,barath2018graph}, as well as globally optimal solvers such as branch and bound (BnB) \cite{li2009consensus,neira2001data}. However, heuristics lack guaranteed global optimality, and their runtime grows exponentially with the outlier ratio, similar to BnB-based solvers. Recently, numerous graph-theoretic methods have emerged to advance consensus maximization in solving problems in robotic vision \cite{enqvist2009optimal,yang2020teaser,shi2021robin,forsgren2024group}. These methods require in-depth analysis of the frontend of the tasks to be solved. This leads to the main contribution of our paper, which is the design of a frontend compatibility test specifically considering the imaging principle and noise model for 2D FLS.

\textbf{Establishing correspondences from 2D FLS observations} is challenging due to well-known issues such as severe noise, nonlinear formation, and a limited field of view (FoV) \cite{teixeira2018multibeam}. Previous work has employed feature point descriptors to establish 2D-2D correspondences, with recent learning-aided methods demonstrating significant progress \cite{li2016utilizing,gode2024sonic}. Graph-theoretic methods have also been explored but are primarily limited to semi-structured environments \cite{santos2019underwater,zhuang2024graph}. In the context of 2D-3D correspondences, research has largely focused on correlating sonar measurements with artificial landmarks \cite{lee2017probability,wang2020acmarker,norman2023actag,pyo2017beam}. This paper focuses on 2D-3D point correspondences, as the geometric characteristics of 3D points facilitate the design of compatibility tests.

\textbf{Outlier rejection in 2D FLS} is often implicitly discussed in the context of data association (e.g., \cite{wang2022virtual,wang2020acmarker,zhuang2024graph}). To the best of the authors’ knowledge, no existing work explicitly focuses on rejecting outliers in 2D-3D point correspondences from 2D FLS observations. In \cite{teixeira2018multibeam}, a series of preprocessing methods are proposed to reject measurement outliers in a single 2D FLS image, but this method targets imaging artifacts rather than outliers in correspondences.

\section{Preliminaries}
\label{sec:preliminaries}
This section introduces the imaging principles of 2D FLS and briefly describes the outlier rejection pipeline.

\subsection{Projection Model of 2D FLS}
\label{subsec:projection}
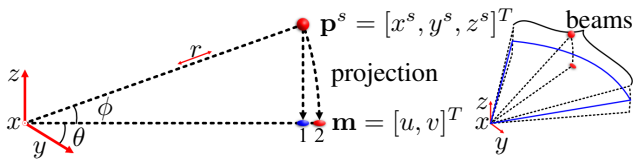
\begin{figure}[h]
	\centering
	\begin{tikzpicture}
		\node[anchor=south west,inner sep=0] (image) at (0,0) 
		{\includegraphics[width=\linewidth]{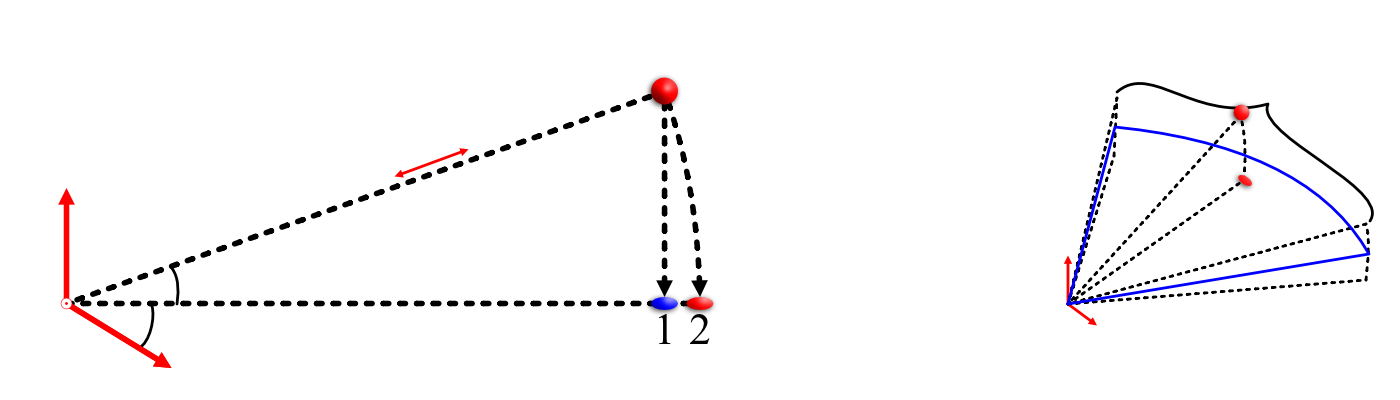}};
		\begin{scope}[shift={(image.south west)}, x={(image.south east)}, y={(image.north west)}]
			\node at (0.4, 1) {1: orthgraphic (approximated); 2: arc projection};
			\node at (0.03,0.24) {$x$};
			\node at (0.03,0.5) {$z$};
			\node at (0.07,0.1) {$y$};
			\node at (0.75,0.24) {$x$};
			\node at (0.75,0.35) {$z$};
			\node at (0.78,0.12) {$y$};
			\node at (0.31,0.63) {$r$};
			\node at (0.6,0.5) {projection};
			\node at (0.93,0.8) {beams};
			\node at (0.17,0.32) {$\phi$};
			\node at (0.13,0.17) {$\theta$};
			\node at (0.65,0.78) {$\mathbf{p}^s = [x^s, y^s, z^s]^T$};
			\node at (0.62,0.25) {$\mathbf{m} = [u, v]^T$};			
		\end{scope}
	\end{tikzpicture}
	\caption{The projection model of 2D FLS. The left part is a side view of one of the dashed sectors in the right part. The arc projection represents the true working mechanism of the sonar.}
	\label{fig:sonar sketch}
\end{figure}
A 3D point $\mathbf{p}^s$ in sonar coordinates can be described using $\left(r, \theta, \phi\right)$ in spherical coordinates:
\begin{equation}
	\mathbf{p}^s = 
	\begin{bmatrix}
		x^s \\
		y^s \\
		z^s
	\end{bmatrix} = 
	\begin{bmatrix}
		r\cos\phi\sin\theta \\
		r\cos\phi\cos\theta \\
		r\sin\phi
	\end{bmatrix},
	\label{equ:spherical-to-cartesian}
\end{equation}
where $r$ is the distance, $\theta$ the bearing angle, $\phi$ the elevation angle. During sonar measurements, $\mathbf{m} = [r,\theta]^T$ are obtained while $\phi$ is lost. When analyzing in the Cartesian coordinate, we abuse the notation for $\mathbf{m}$ and have
\begin{equation}
	\mathbf{m} = 
	\begin{bmatrix}u \\ v\end{bmatrix}=
	\begin{bmatrix}r\sin\theta \\ r\cos\theta\end{bmatrix}=
	\begin{bmatrix}\cos^{-1}\phi&0&0\\0&\cos^{-1}\phi&0 \end{bmatrix}
	\mathbf{p}^s.
	\label{equ:projection}
\end{equation}
Eq. \eqref{equ:projection} demonstrates the projection process of 2D FLS. However, it is often the case that we only know $\mathbf{p}^w_i$, and thus Eq. \eqref{equ:projection} becomes
\begin{equation}
	\mathbf{m} = 
	\begin{bmatrix}\cos^{-1}\phi&0&0\\0&\cos^{-1}\phi&0 \end{bmatrix}
	\big(\mathbf{R}^s_w\mathbf{p}^w + \mathbf{t}^s_w\big),
	\label{equ:projection with world coordinate}
\end{equation}
where $\mathbf{R}^s_w \in SO(3)$ and $\mathbf{t}^s_w \in \mathbb{R}^{3}$ are the unknown transformation parameters. The typical range of $r$ is 120 m, but it is often limited to $\sim$ 20 m for standard surveying tasks and to $\le$ 5 m for high-resolution perception. For $(\theta, \phi)$, i.e. the FoV of 2D FLS, the typical ranges are $|\theta| \le 65^\circ$ and $|\phi| \le 10^\circ$. We use $\{\text{max, min}\}$ as subscripts to denote the symmetric bounds. Since $\phi_\text{max}$ is small, $\cos^{-1}\phi$ in Eq. \eqref{equ:projection} can be approximated as $1$.  This linearization technique results in an orthographic approximation, which is often adopted for computational convenience \cite{su2024analysis, hurtos2015fourier}. The projection process and the orthographic approximation are illustrated in Fig. \ref{fig:sonar sketch}.

For noise modeling, we assume additive Gaussian noise on the raw data, which is a common practice for range-bearing sensors \cite{xu2025incorporating,norman2023actag}. We denote the noisy measurements as follows: $r^* = r + \eta$, where $\eta \sim \mathcal{N}(0, \sigma_r^2)$, and $\theta^* = \theta + \epsilon$, where $\epsilon \sim \mathcal{N}(0, \sigma_\theta^2)$. In the subsequent discussion, we will explicitly use the superscript $*$ to distinguish the influence of noise on $\mathbf{m}$ when it is mentioned.

\subsection{Outlier Rejection Pipeline}
\begin{figure}
	\includegraphics[width=\linewidth]{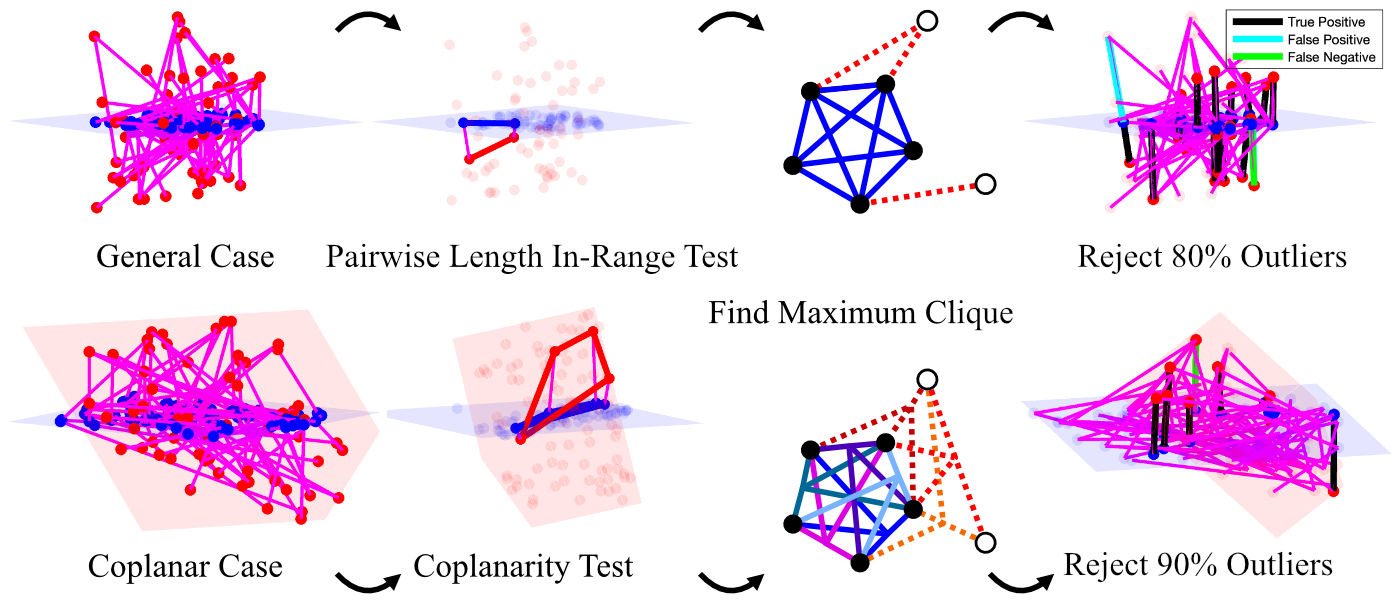}
	\caption{A schematic diagram of the proposed method.}
	\label{fig:beginner}
\end{figure}
Suppose we already have an initial guess of correspondences $\mathcal{C}_{\text{init}} \vcentcolon= \{(\mathbf{p}^w_i, \mathbf{m}_i) \mid i=1,\dots,N\}$. The goal is to estimate the outliers \(\mathcal{O}_\text{est}\) in such a way that it closely approximates the ground truth \(\mathcal{O}_\text{gt}\). The algorithm is sumarized in Fig. \ref{fig:beginner}. First, we pass the correspondences through compatibility tests to evaluate their compatibility. Next, we build a compatibility graph, using the correspondences as vertices. We connect groups of sufficiently compatible measurements with edges, or hyperedges if the group size exceeds two. Finally, we identify the maximum clique of the graph and classify the nodes within this clique as inliers. These nodes are mutually connected, demonstrating mutual compatibility, and together form the largest such group. This type of pipeline is widely adopted in the literature, with recent remarkable advancements that consider compatibility tests involving more than two measurements, as represented in \cite{shi2021robin,forsgren2024group}. The main difference compared to these works is that we propose specific compatibility tests tailored for 2D FLS. The tests are notably distinct due to the unique noise model and imaging principles of 2D FLS, making its design non-trivial.

\section{Compatibility Tests}
\label{sec:compatibility tests}
This section presents two compatibility tests that are specifically designed considering the imaging principles of 2D FLS. The first one is the pairwise length in-range test, referred to as the \texttt{in-range test}, which is based on the length constraint and is suitable for general cases. The second one is the \texttt{coplanarity test}, which is exclusively applicable to cases where the 3D points are coplanar but offers greater robustness.

\subsection{Pairwise Length In-Range Test}
\label{subsec:in-range}
Given $\mathcal{C}_\text{init}$, we can partition it into two subsets as $\mathcal{P} \vcentcolon= \{\mathbf{p}^w_i\}^N_{i=1}$ and $\mathcal{M} \vcentcolon= \{\mathbf{m}_i\}^N_{i=1}$. By forming undirected edges pairwise between the elements in $\mathcal{P}$, we obtain the edge set $\mathcal{EP} \vcentcolon= \{\vv{\mathbf{p}^w_{ij}}\mid i < j \le N\}$. Considering the Eq. \eqref{equ:projection with world coordinate} and the FoV of 2D FLS, we have Proposition \ref{pro:length constraint}.
\begin{proposition}[Length Constraint]
	\label{pro:length constraint}
	Given two pairs of $(\mathbf{p}^w_i, \mathbf{m}_i)$ and $(\mathbf{p}^w_j, \mathbf{m}_j)$, if they are both inliers, the $\| \vv{\mathbf{p}^w_{ij}} \|$ should lie within the feasible region $\mathcal{F}$ defined by $\mathbf{m}_{\{1,2\}}$:
	\begin{multline}
		\mathcal{F} = \biggl \{x \in \mathbb{R} \big| \big(r_i^2 + r_j^2 - 2r_i r_j \mathbf{X}_{\text{min}}\big)^{\frac{1}{2}} \le x  \le \hfill
		\\
		\hfill \big(r_i^2 + r_j^2 - 2r_i r_j \mathbf{X}_{\text{max}}\big)^{\frac{1}{2}} \biggr \},
		\\
		\mathbf{X}_{\text{min}} = \big(\cos(\theta_i-\theta_j)-1\big)\cos^2\phi_\text{max}+1 \hfill
		\\
		\mathbf{X}_{\text{max}} = \cos(\theta_i-\theta_j) - \big(1+\cos(\theta_i-\theta_j)\big)\sin^2{\phi_\text{max}} \hfill
		\label{equ:length bounds}
	\end{multline}
	where $r_{\{i,j\}}$ and $\theta_{\{i,j\}}$ are distance and bearing angle in $\mathbf{m}_{\{i,j\}}$. The upper bound is reached when the 3D points are on opposite sides of the $\phi$ bound, while the lower bound is reached when they are on the same side. In the special case where $\theta_i = \theta_j$, the points only need to be on the same $\phi$ to reach the lower bound.
\end{proposition}

A proof is provided in the Appendix A. Using Proposition \ref{pro:length constraint}, we can evaluate each element in \(\mathcal{EP}\). If an element passes the test, meaning its length falls within \(\mathcal{F}\), we establish edges between the corresponding vertices, thereby forming a compatibility graph. The philosophy behind the design of length constraint is to bound the uncertainty caused by the missing $\phi$ during sonar measurements. Fortunately, because the $\phi_\text{max}$ is always small, one can check through Eq. \eqref{equ:length bounds} that the resulting $\mathcal{F}$ is also small, making the length constraint an efficient tool to filter out overly long or short edges.

However, Proposition \ref{pro:length constraint} does not account for noisy measurements. As a result, the derived bounds, which represent the uncertainty, are overly confident when noise is present and may significantly reject inliers, which is undesired. To accommodate noisy $\mathbf{m}^*$, the bounds in Eq. \eqref{equ:length bounds} should be expanded for greater tolerance. We denote $\mathcal{F}^*$ as the feasible region after considering the influence of noise. Since the result does not have a simple form, we provide the detailed derivation in the Appendix B.

\subsection{Coplanarity Test}
\label{subsec:coplanarity test}
In this section, we derive the \texttt{coplanarity test} based on the approximation that the imaging process of 2D FLS can be considered as orthographic projection (Sec. \ref{subsec:projection} and Fig. \ref{fig:sonar sketch}). It is well-known that the orthographic projection process of a 3D plane can be represented as a 2D affine transformation in the viewing plane\footnote{Considering the projection in the camera coordinate, the $z$ component is omitted, resulting precisely in a 2D affine transformation.}. In the context of coplanarity verification, this implies that for any $4$-tuple selected from $\mathcal{C}_\text{init}$, if $\left\{(\mathbf{p}^w_i, \mathbf{m}_i) \mid i = 1,2,3,4\right\}$ are all correct correspondences, the 2D affine matrix estimated from any three correspondences in it should consistently apply to the fourth one. First, we consider the case of estimating the 2D affine matrix $\mathbf{A}$ from $i=\{1,2,3\}$, and have
\begin{equation}
	\mathbf{A}
	\underbrace{
		\begin{bmatrix}
			\mathbf{p}^w_1 & \mathbf{p}^w_2 & \mathbf{p}^w_3 \\
			1 & 1 & 1
		\end{bmatrix}}_{\mathbf{P}}
	= 
	\underbrace{
		\begin{bmatrix}
			\mathbf{m}_1 & \mathbf{m}_2 & \mathbf{m}_3
		\end{bmatrix}}_{\mathbf{M}},
\end{equation}
where $\mathbf{A}$ has a dimension of $2\times 3$. If the points in $\mathbf{P}$ are noncollinear, $\mathbf{A}$ can be determined through $\mathbf{A} = \mathbf{M}\mathbf{P}^{-1}$. Then, we evaluate $\mathbf{A}$ on the fourth point:
\begin{equation}
	\mathbf{A}
	\begin{bmatrix}
		\mathbf{p}^w_4 \\ 1
	\end{bmatrix}-
	\mathbf{m}_4
	=
	\mathbf{M}
	\underbrace{
		\mathbf{P}^{-1}
		\begin{bmatrix}
			\mathbf{p}^w_4 \\ 1
		\end{bmatrix}
	}_{\mathbf{b}=[b_1,b_2,b_3]^T} -
	\mathbf{m}_4	
	= \mathbf{Mb}-\mathbf{m}_4.
	\label{equ:coplanarity check}
\end{equation}
The linearity in Eq. \eqref{equ:coplanarity check} facilitates the modeling of the residuals' distribution under the assumption that the raw measurements are affected by Gaussian-distributed noise, as stated in Sec. \ref{subsec:projection}. Through derivation, we find that the residuals' distribution can also be approximated as Gaussian. Specifically, if all four correspondences are correct, the distribution becomes zero-mean. The detailed derivation is provided in Appendix C. After dividing the residuals by the calculated standard deviation and considering the sum of their squares, the results would follow a $\mathcal{X}^2_2$-distribution, where the subscript indicates that it has two degrees of freedom. For the \(\tbinom{4}{3}\) combinations, we obtain four $\mathcal{X}^2$ values, the sum of these values follows a $\mathcal{X}^2_8$-distribution. By selecting a specific \textit{p}-value (e.g., 0.01) for thresholding, we can reject $4$-tuples that fail the \texttt{coplanarity test}. Finally, if a $4$-tuple passes the test, we form a size $4$ hyperedge between the involved vertices (correspondences). By iterating through all $4$-tuples, we complete the construction of the compatibility graph, which is a $4$-uniform hypergraph.

The above discussion has outlined the steps for the \texttt{coplanarity test} and the method for constructing the compatibility graph. Now, we turn to a more detailed discussion on \textbf{modeling the noise in $r^*$}. To ensure the orthographic assumption holds, the distance measurement should be adjusted to $r\cos\phi$ to accurately represent the $(x^s, y^s)$ of the 3D point. Thus, the probability density function (PDF) of the observed $r^*$ becomes
\begin{equation}
	\label{equ:pdf of r}
	p(r^*\mid\phi,r) = \frac{1}{\sigma_r\sqrt{2\pi}}\exp\left(-\frac{(r^*-r\cos\phi)^2}{2\sigma_r^2}\right).
\end{equation}
$\phi$ cannot be obtained as prior knowledge and is generally assumed to follow a uniform distribution within the elevation FoV, i.e. $\phi \sim U(\phi_{\text{min}},\phi_{\text{max}})$. To derive the actual PDF, we should marginalize over $\phi$. While obtaining a closed-form PDF is challenging, we approximate it using another Gaussian distribution. Qualitative results in Sec. \ref{subsec:experiments on approx r} will demonstrate that this approximation is effective in practical applications. The first- and second-order moments of the estimated PDF are 
\begin{align}
	\mu_{\text{est}} &= \frac{r\sin\phi_\text{max}}{\phi_\text{max}},
	\\
	\sigma_{\text{est}}^2 &= \sigma_r^2 + \mu_{\text{est}}^2\left(\frac{1}{2}+\frac{\sin2\phi_\text{max}}{4\phi_{\text{max}}} - \left(\frac{\sin\phi_{\text{max}}}{\phi_{\text{max}}}\right)^2\right).
\end{align}
These results can be trivially derived using the law of total expectation and the law of total variance \cite{chung2000course}.

\section{Experiments and Results}
\label{sec:experiments and results}

\subsection{General Cases}
\label{subsec:general cases}
\begin{figure*}
	\centering
	\includegraphics[width=0.32\linewidth]{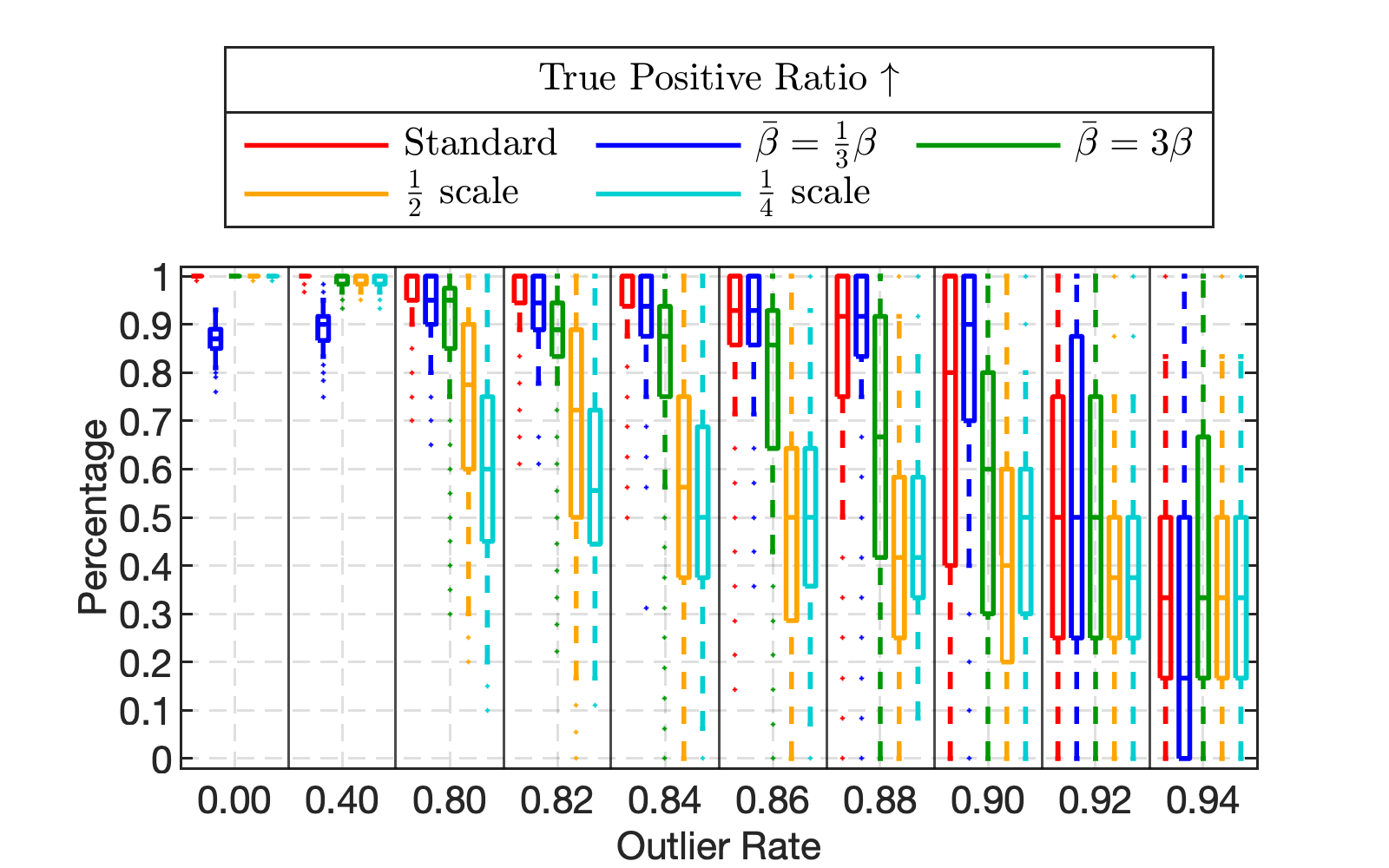}
	\includegraphics[width=0.32\linewidth]{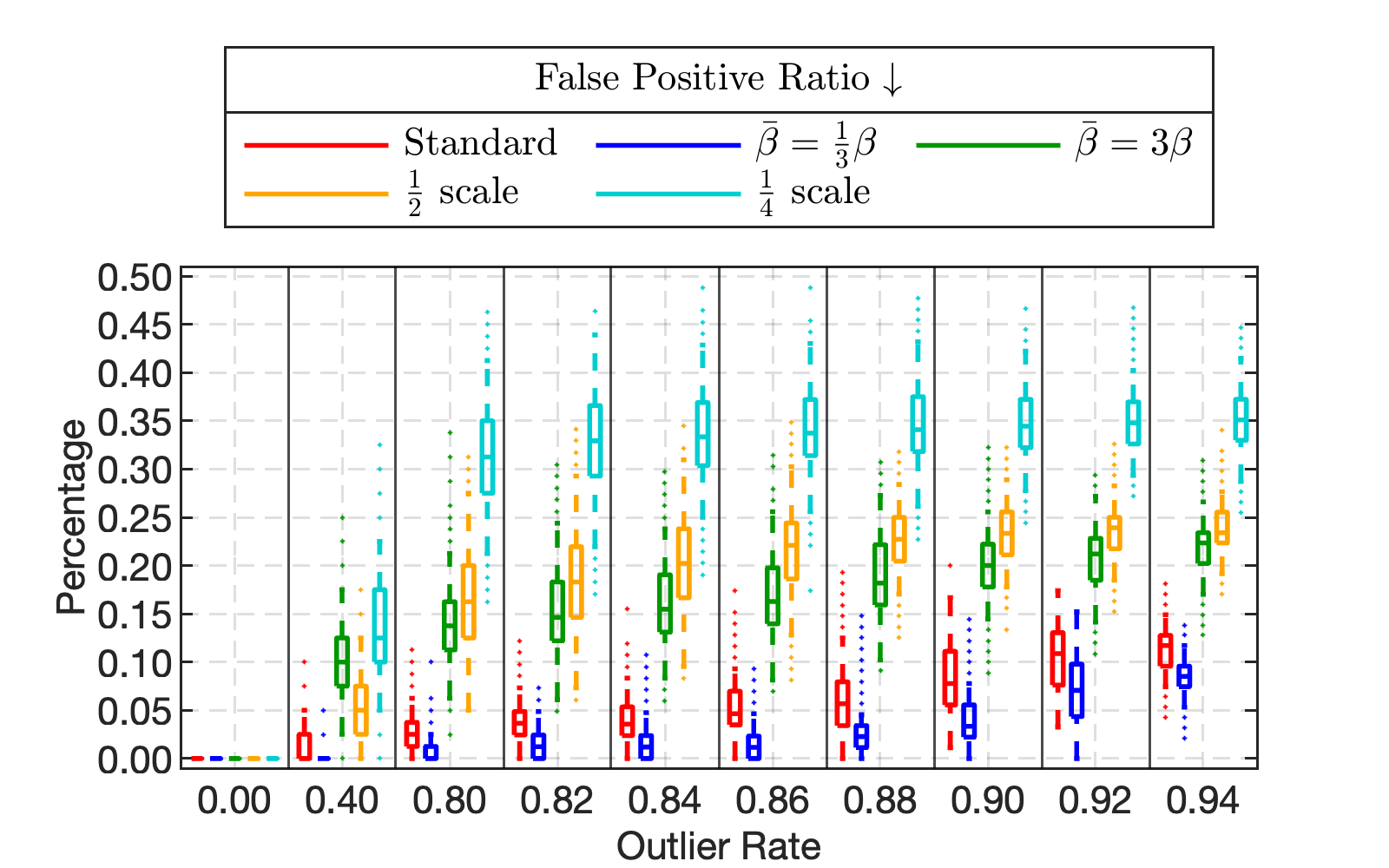}
	\includegraphics[width=0.32\linewidth]{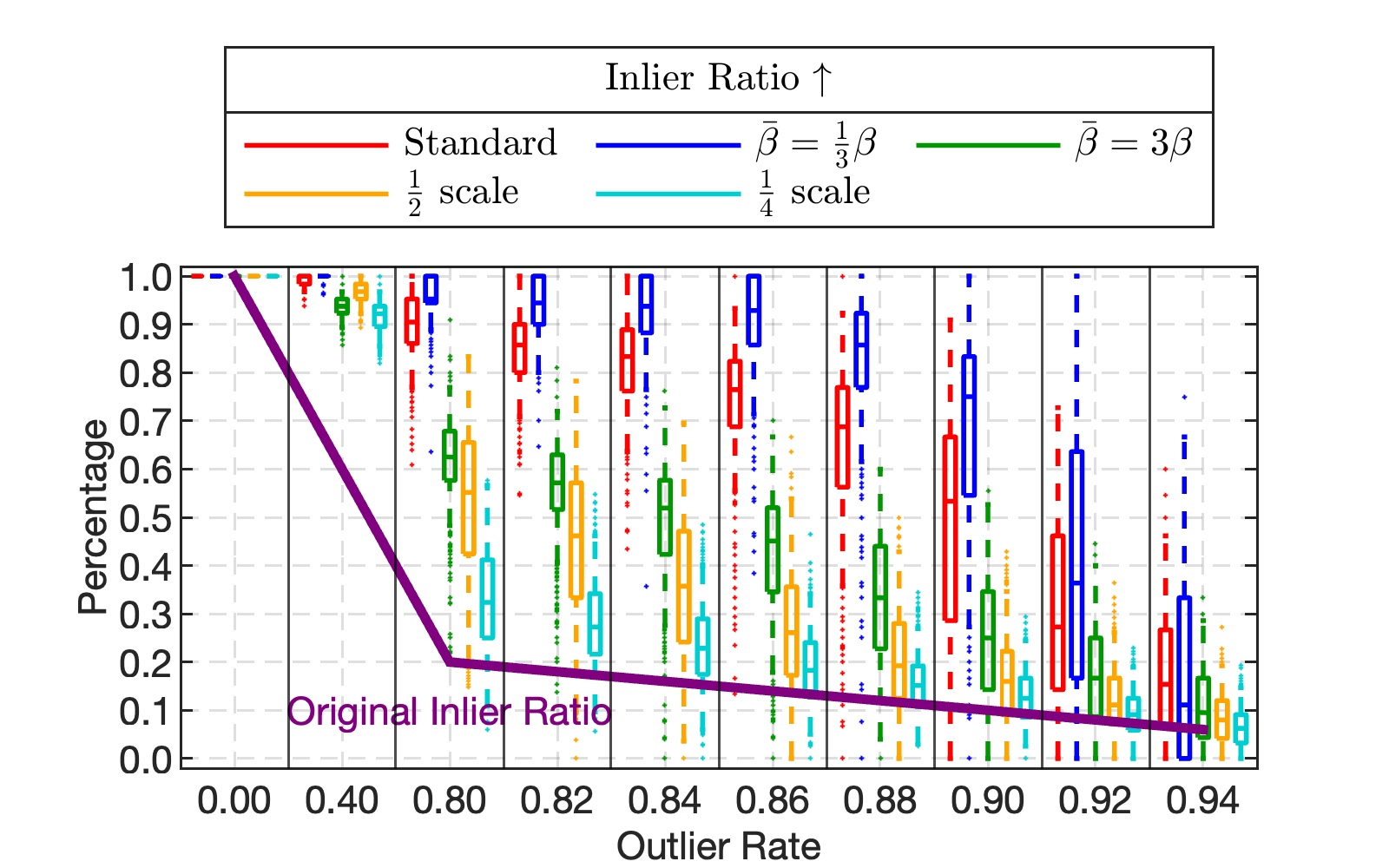}
	\caption{The results of the general cases with different parameter settings.}
	\label{fig:general cases}
\end{figure*}
For general cases, the \texttt{in-range test} and the simple graph maximum clique solver are used. The \texttt{in-range test} is implemented in MATLAB R2024b, leveraging vectorization to speed up the process. For the simple graph maximum clique solver, we utilize the one reported in \cite{rossi2015parallel}. In the sonar coordinate system, 100 3D points were randomly generated within a box defined by the ranges \([-0.6, 0.6] \times [1.6, 2.8] \times [-0.3, 0.3]\). Using Eq. \eqref{equ:projection}, we obtained \(\mathcal{M}\), which is influenced by additive Gaussian noise. Subsequently, random 3D transformations were generated and applied to these 3D points to obtain \(\mathcal{P}\). By pairing the elements of \(\mathcal{M}\) and \(\mathcal{P}\), we established the ground truth correspondences \(\mathcal{C}_\text{gt}\). To simulate different outlier ratios, a certain proportion of the true correspondences were replaced with randomly generated ones between arbitrary \(\mathbf{p}^w\) and \(\mathbf{m}\).  For the 2D FLS-specific parameters, we set \(\phi_\text{max} = 7^\circ\), \(\sigma_r = 0.005 \, \text{m}\), and \(\sigma_\theta = 0.5^\circ\). These values are consistent with typical real-world sonar specifications. For the control group \textbf{Standard}, we defined \(\beta_{r} = 3\sigma_r\) and \(\beta_\theta = 3\sigma_\theta\), assuming accurate noise estimation. For comparison, we established four groups: (1) \textbf{Reduced Bound}: We set \(\bar{\beta} = \frac{1}{3}\beta\) to simulate inaccurate noise estimation. (2) \textbf{Expanded Bound}: We set \(\bar{\beta} = 3\beta\). (3) \textbf{Half Scale}: We reduced the box size for 3D point generation to \(\frac{1}{2}\) (resulting in \(\frac{1}{8}\) of the original volume). This implies that more false correspondences may fall within \(\mathcal{F}^*\). (4) \textbf{Quarter Scale}: We reduced the scale by a factor of \(\frac{1}{4}\). The resulting box size is a cube measuring \(0.3 \, \text{m} \times 0.3 \, \text{m} \times 0.15 \, \text{m}\), which is close to the size of the acoustic fiducial markers \cite{wang2020acmarker,norman2023actag}.  For quantitative evaluation, we employ the following metrics: true positive ratio (TPR), false positive ratio (FPR), and inlier ratio (IR, also referred to as precision). For each group and each selected outlier ratio, we conducted 500 trials and recorded the results.

The results are presented in Fig. \ref{fig:general cases}, and the discussion is as follows:  
(1) We observe that \textbf{Standard} performs well when the outlier ratio is not extreme (\(\leq 80\%\)). At an 80\% outlier ratio, the mean and median values of IR are 88.61\% and 90.48\%, respectively. As the outlier ratio increases, the performance declines. Up to 94\%, IR shows minimal improvement. Nevertheless, the size of the correspondences set is significantly reduced, which benefits the minimal solver in finding a better initial solution. (2) The \textbf{Reduced Bound} can reject more false correspondences; however, this strictness may also lead to the rejection of additional true correspondences. As a result, while the IR remains relatively high, the number of true correspondences decreases. On the other hand, the \textbf{Expanded Bound} does not retain more true correspondences, as it allows too many false ones to pass the \texttt{in-range test}, making it more likely that the identified maximum clique is not the true inlier set. Nevertheless, even under such a conservative noise estimation, at an 80\% outlier ratio, the mean and median values of IR are 62.16\% and 62.50\%, respectively, representing a 3x improvement compared to the original inlier ratio. (3) The results of \textbf{Half Scale} and \textbf{Quarter Scale} demonstrate acceptable performance at low outlier ratios but exhibit significant performance degradation as the outlier ratio increases. This outcome is expected because such configurations cannot effectively utilize the length constraint to filter out false correspondences. However, such scenarios are relatively uncommon in practice, as the establishment of 2D-3D point correspondences typically involves the entire scene, e.g., tracking and relocalization in SLAM \cite{mur2015orb}, which prevents correspondences from being confined to a small space. Even in the case of recognizing fiducial markers \cite{wang2020acmarker,norman2023actag}, while the 3D points may be concentrated in a specific region, the corresponding $\mathbf{m}$ on the other end of the correspondences are likely distributed across the entire FoV.

In summary, although some groups achieve satisfactory outlier rejection performance even at a 90\% outlier ratio, considering the method's susceptibility to various influencing factors, the results at an 80\% outlier ratio demonstrate overall strong performance. Therefore, we conservatively conclude that the outlier rejection method in general cases is robust to an 80\% outlier ratio.

\subsection{Coplanar Cases}
\begin{figure*}
	\centering
	\includegraphics[width=0.32\linewidth]{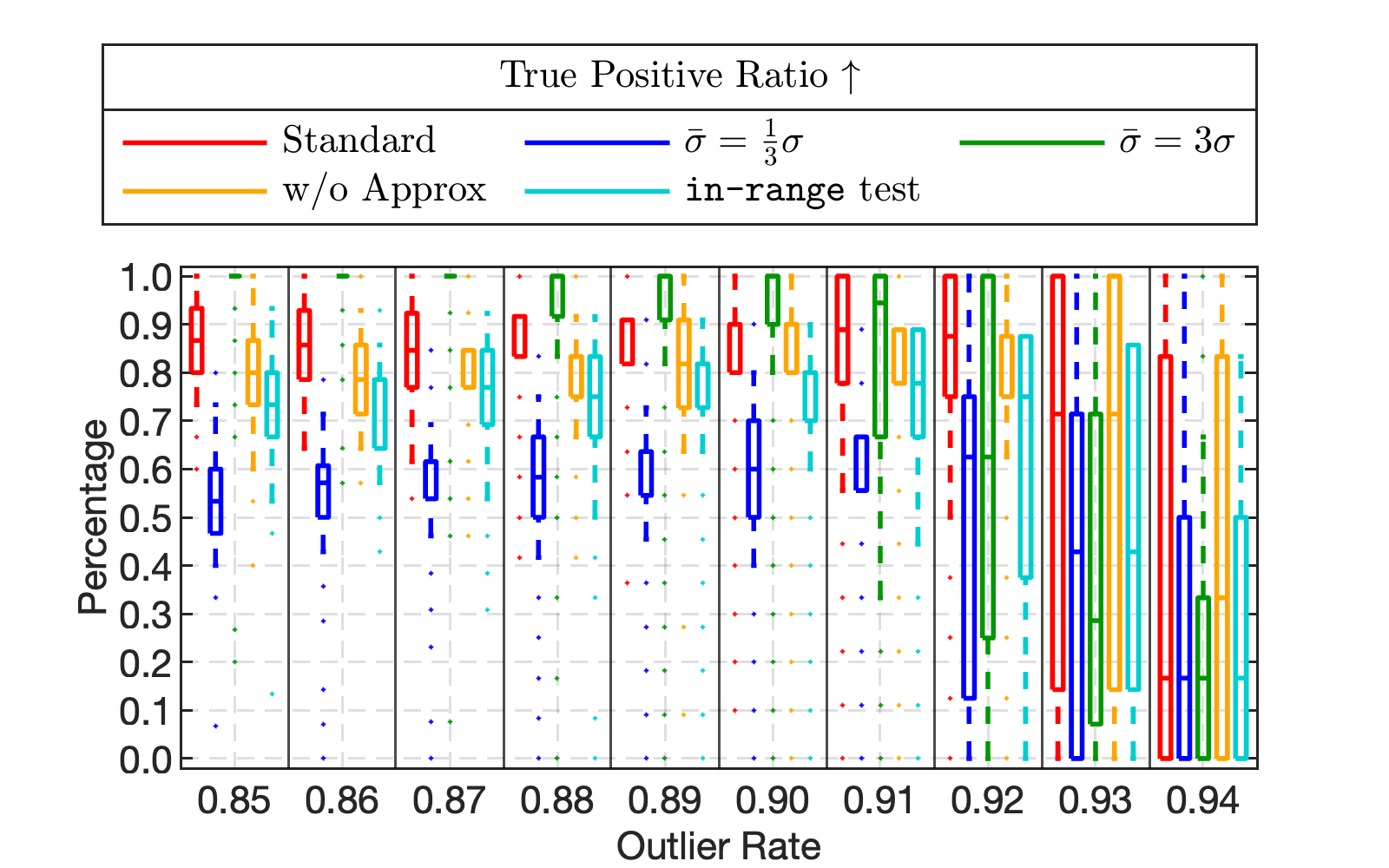}
	\includegraphics[width=0.32\linewidth]{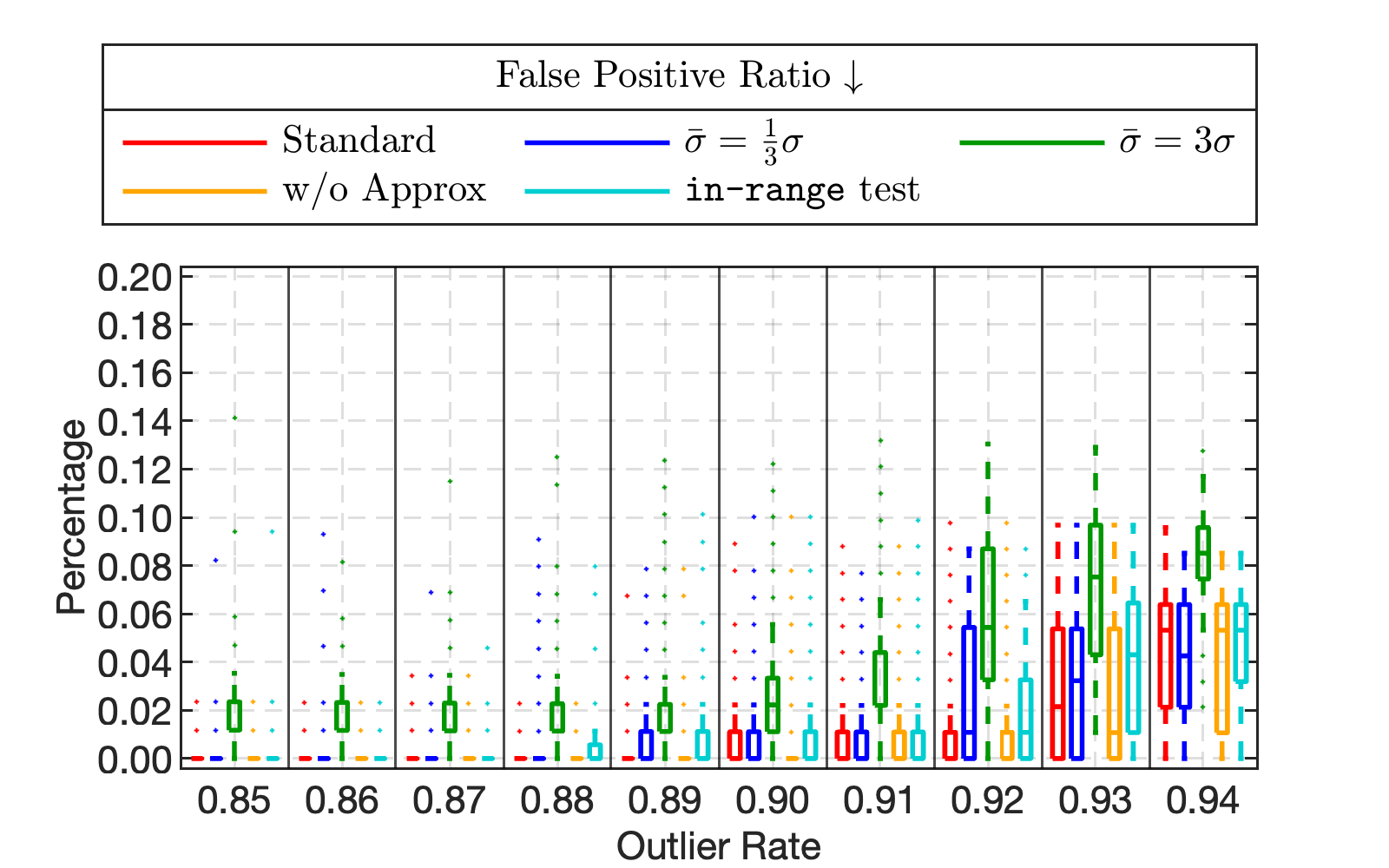}
	\includegraphics[width=0.32\linewidth]{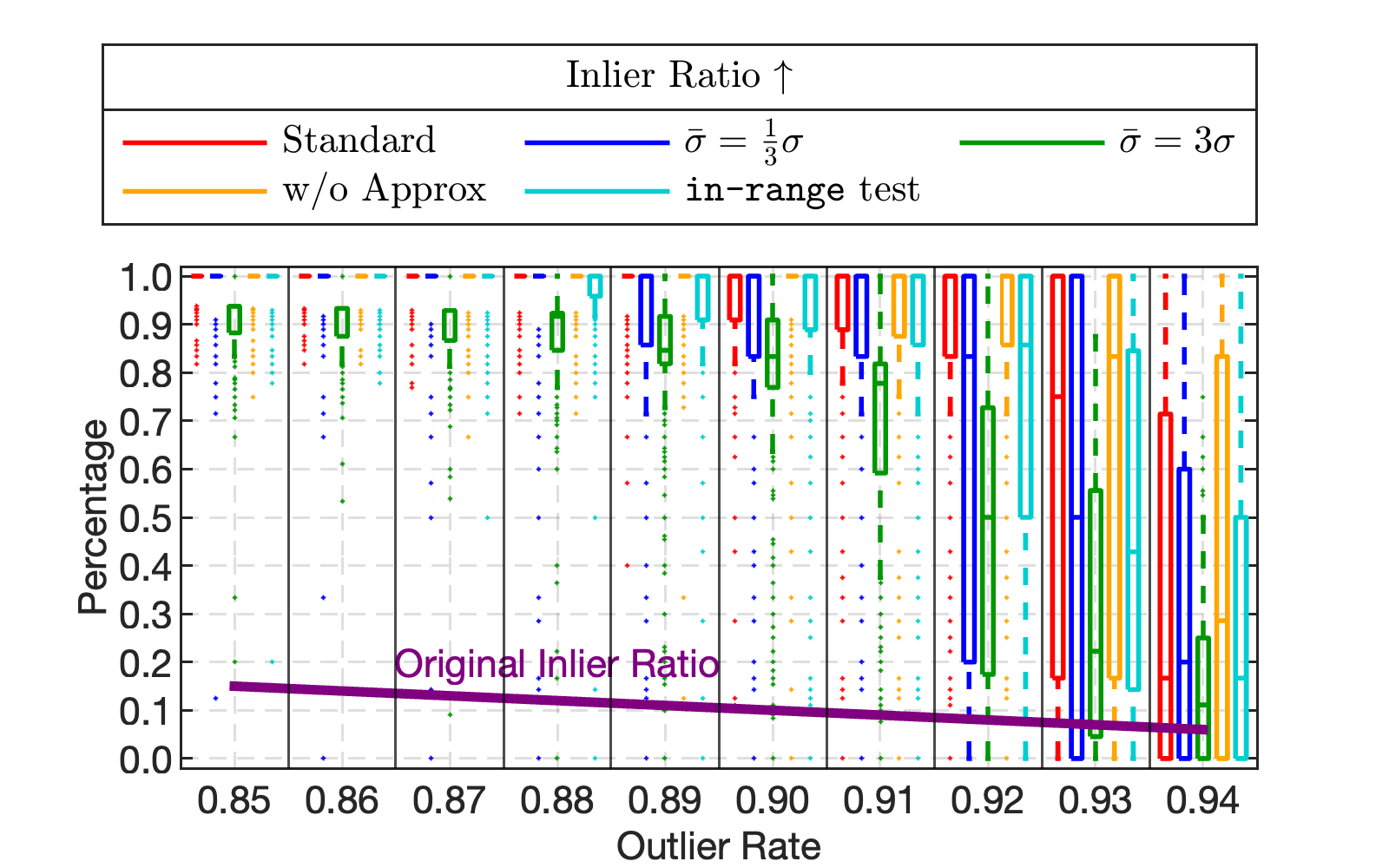}
	\caption{The results of the coplanar cases with different parameter settings.}
	\label{fig:coplanar cases}
\end{figure*}
For coplanar cases, the \texttt{coplanarity test} and the hypergraph maximum clique solver are used. The \texttt{coplanarity test} is implemented in C++ with multi-threading (8 threads in experiments). For the hypergraph maximum clique solver, we adopt the heuristic solver proposed in \cite{forsgren2022group}, which sacrifices a certain degree of accuracy but provides a significant speedup. We have adopted the simulation settings from Sec. \ref{subsec:general cases}, with the addition of forcing the plane to pass through a point where $z^s = 0$, and both $x^s$ and $y^s$ are within the bounds $[-0.15, 0.15] \times [2.05, 2.35]$. Furthermore, we constrain the angle between the plane and the $xy$-plane to range between $5^\circ$ and $70^\circ$, ensuring that the coplanar points are clearly observable. For the control group \textbf{Standard}, we assume accurate noise estimation. For comparison, we established four groups: (1) \textbf{Underestimated}: We set the estimated standard deviation of both \(r\) and \(\theta\) to one-third of the ground truth. (2) \textbf{Overestimated}: We set it to three times the ground truth. (3) \textbf{No Approx}: We did not adopt the approximation of \(r\) as described in Sec. \ref{subsec:coplanarity test}. (4) \textbf{With In-Range}: We incorporated the \texttt{in-range test} into the algorithm, specifically rejecting any \(4\)-tuple that passes the \texttt{coplanarity test} if any of the two pairs fails the \texttt{in-range test}.

The results are presented in Fig. \ref{fig:coplanar cases}, and the discussion is as follows: (1) The proposed method in coplanar cases is more robust to high outlier ratios compared to general cases. The IR stays close to 100\%, starting to decrease only after reaching 90\%. (2) \textbf{Underestimated} shows a significant decline in TPR, but still maintains a very high IR. \textbf{Overestimated} shows a slight decline in FPR, resulting in the IR stabilizing around 90\%. Although not as good as other groups, there is still a substantial improvement in the original inlier ratio. Based on these results, we suggest that a conservative noise estimation is preferred. (3) \textbf{No Approx} demonstrates comparable IR performance but exhibits a slight decline in TPR, resulting in fewer included inliers. This validates the effectiveness of considering the missing $\phi$ and the approximation of $r$. (4) \textbf{With In-Range} shows decreased performance compared to \textbf{Standard}, indicating the need to explore a softer integration of the two methods to better leverage their combined effects.

In summary, considering both the performance degradation trend and the improvement relative to the original inlier ratio, we conclude that the outlier rejection method in coplanar cases is robust to an outlier ratio of up to 90\%.

\subsection{Evaluating the Approximation of $r$}
\label{subsec:experiments on approx r}
\begin{figure}[h]
	\centering
	\includegraphics[width=0.32\linewidth]{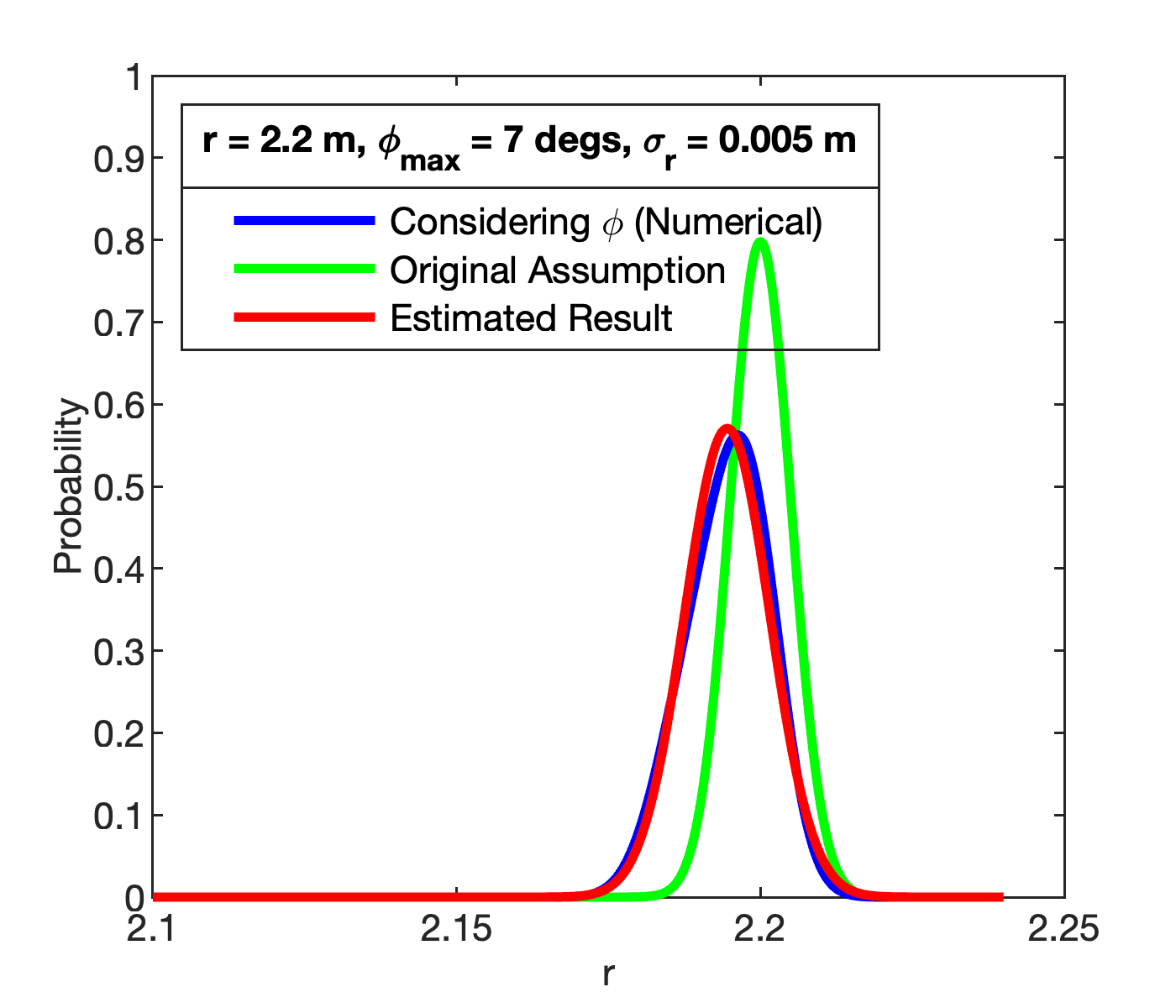}
	\includegraphics[width=0.32\linewidth]{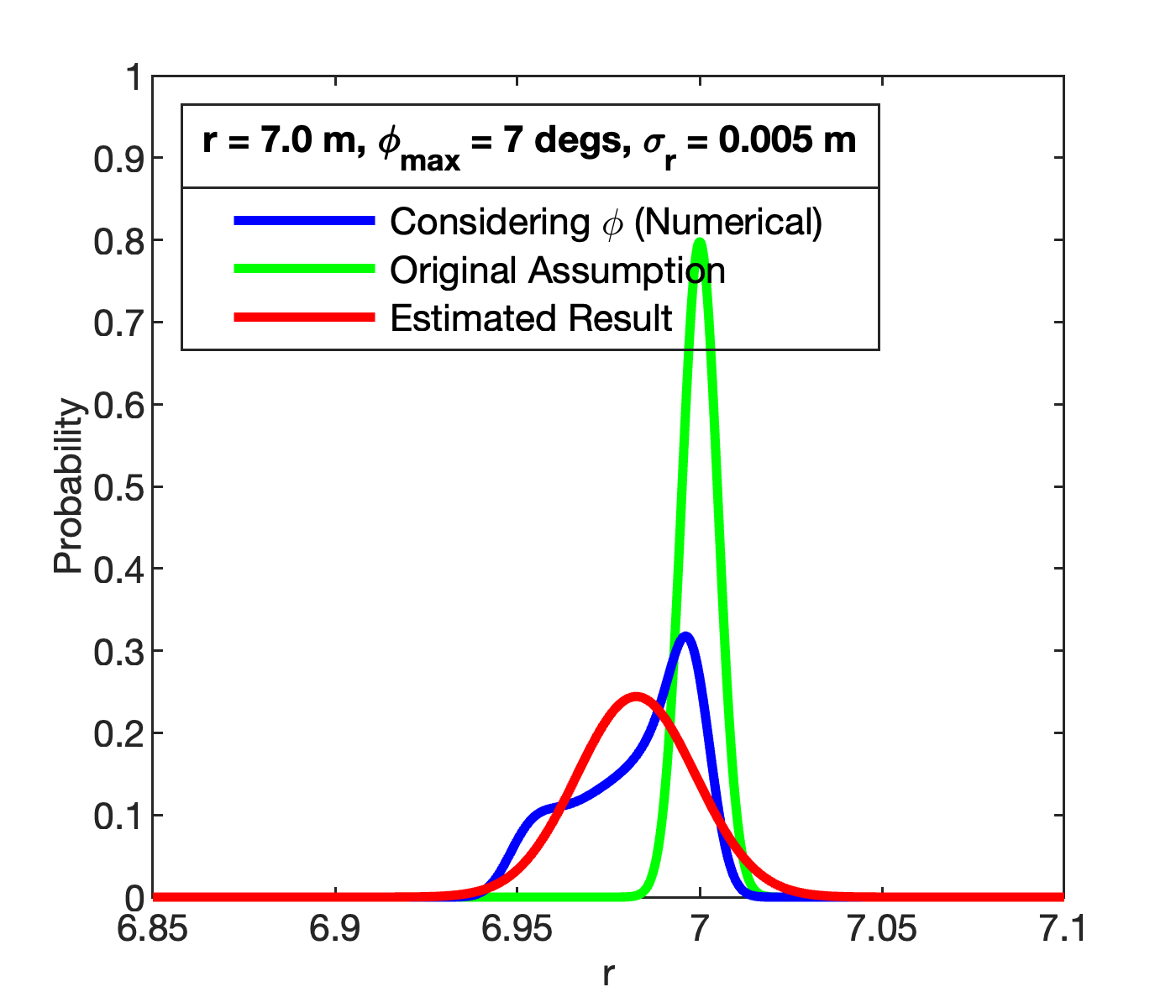}
	\includegraphics[width=0.32\linewidth]{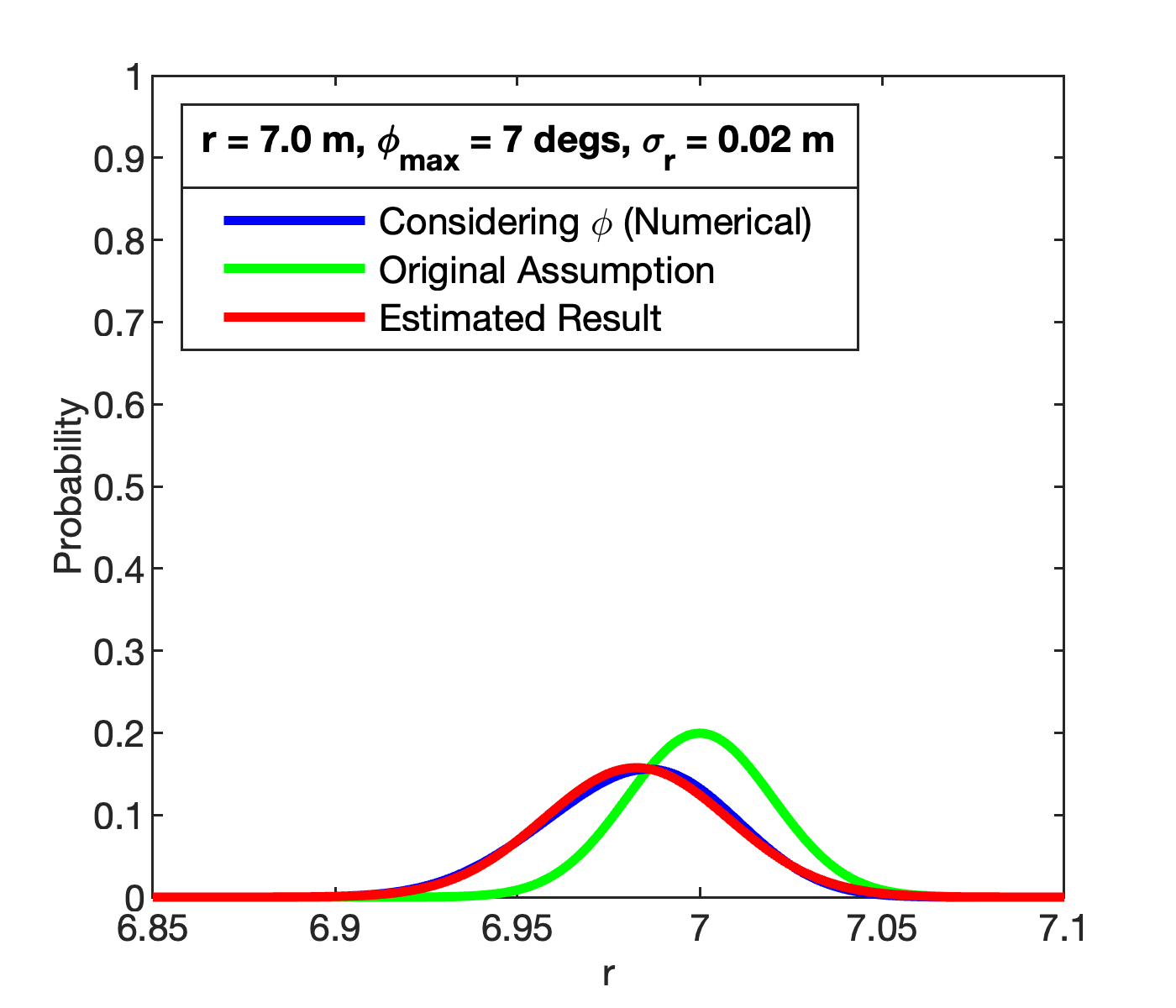}
	\\
	\includegraphics[width=0.32\linewidth]{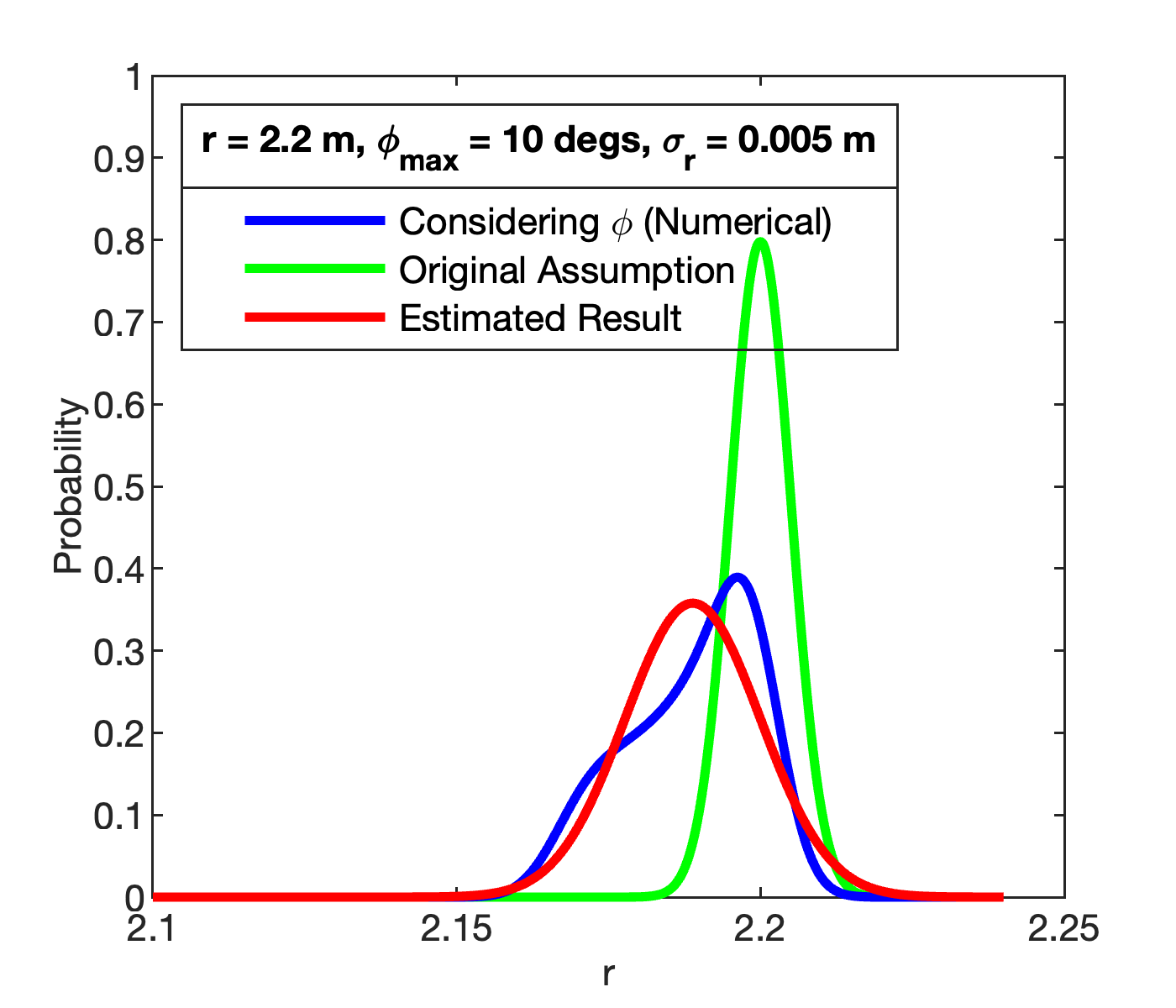}
	\includegraphics[width=0.32\linewidth]{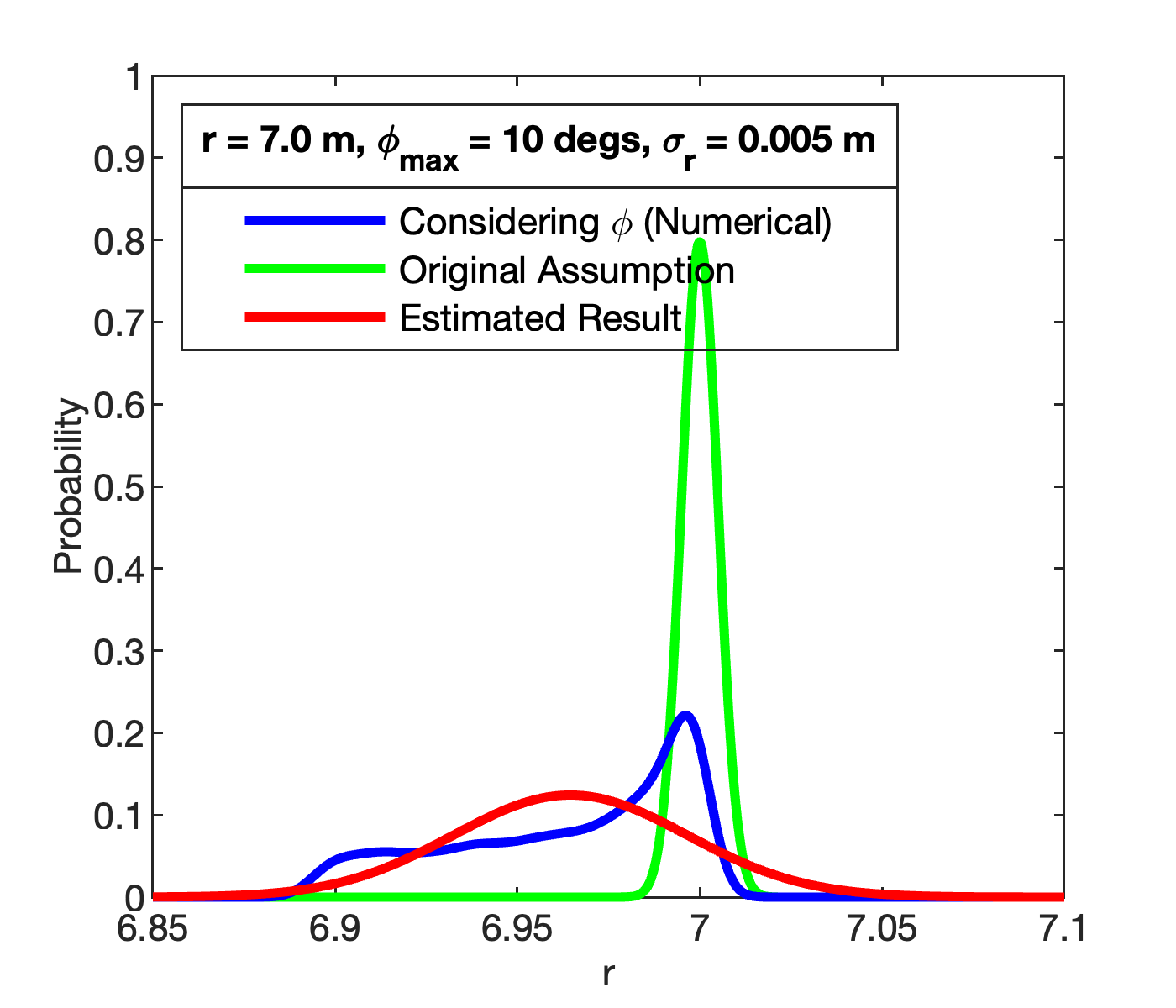}
	\includegraphics[width=0.32\linewidth]{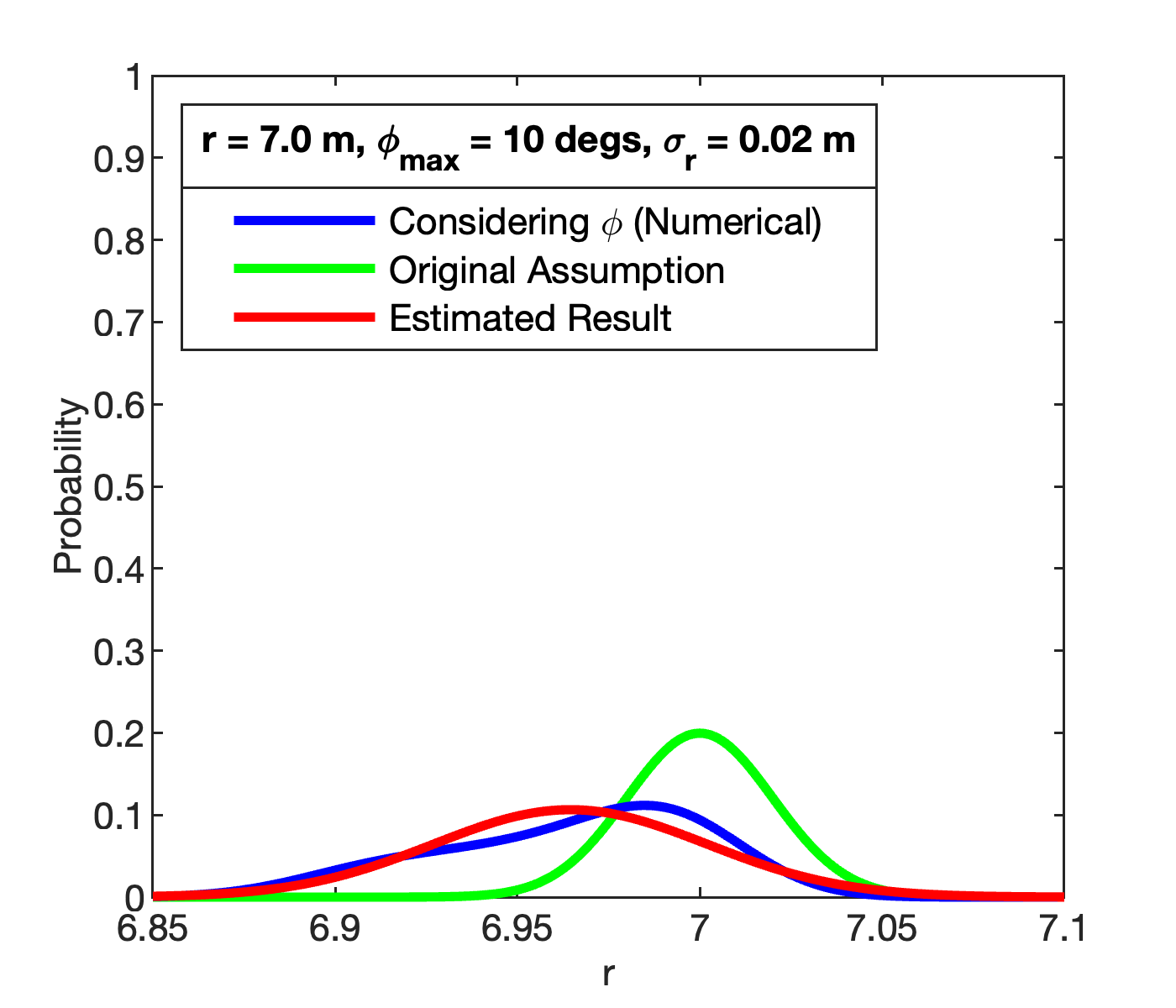}
	\caption{The results of evaluating the approximation effect under different $r$, $\sigma_r$ and $\phi_{\text{max}}$. The closer the red and blue curves, the better the approximation effect.}
	\label{fig:r_distribution}
\end{figure}
We evaluate the approximation of the distribution of $r$ under the orthographic assumption, varying distances (2.2 m and 7 m), $\sigma_r$ (0.005 m and 0.02 m), and $\phi_{\text{max}}$ ($7^\circ$ and $10^\circ$). It should be noted that the value used in the simulation is reasonable, which can be verified by any 2D FLS manuals. The results are illustrated in Fig. \ref{fig:r_distribution}. It shows that in practical use at medium range, e.g., 2.2 m, the approximation is good at small $\phi_\text{max}$ and is acceptable at large $\phi_\text{max}$. When $r$ increases, the approximation gradually becomes less accurate. However, if $\sigma_r$ is also increased, the approximation improves again. Based on the results and analysis, we conclude that our approximation is effective. The increase in \( \sigma_r \) is a reasonable assumption due to several factors, such as errors in measuring the speed of sound, the influence of multipath reverberation \cite{urick1983principles}, and the fact that as angular uncertainty grows with distance, the uncertainty in the measured data also increases, thereby amplifying the uncertainty in feature point extraction.

\subsection{Time Cost}
\begin{figure}
	\centering
	\includegraphics[width=0.7\linewidth]{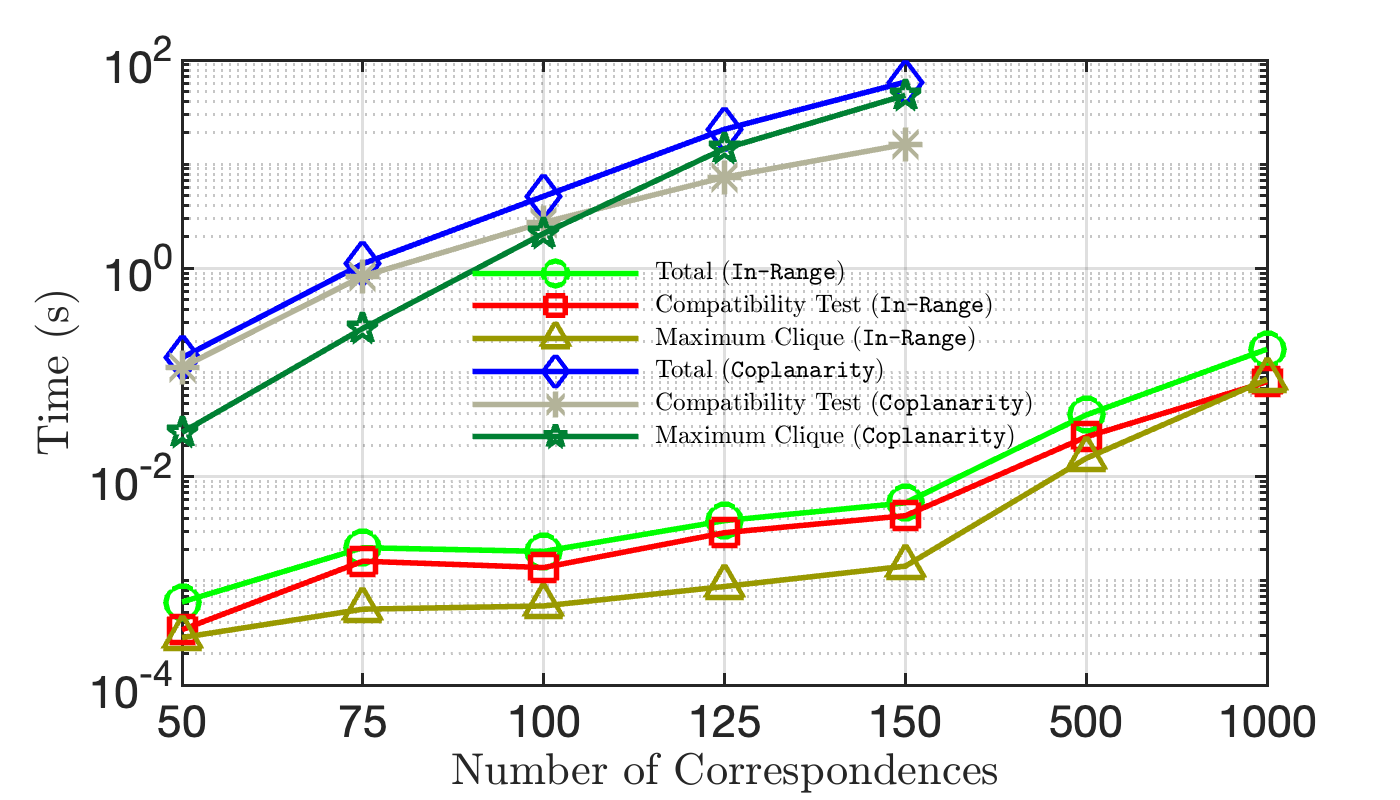}
	\caption{The results of the time cost evaluation.}
	\label{fig:time cost}
\end{figure}
We set the outlier ratio to 80\% and conducted 100 trials for varying numbers of correspondences to evaluate the speed of the proposed outlier rejection method. Both general cases and coplanar cases were assessed. In addition to the average running time for one trial, we also report the time cost for compatibility tests and the maximum clique solver (the graph construction is trivial). The simulation was run on a MacBook Air M1 \footnote{Reported similar CPU performance to the Intel Core i5-1245U.\\https://nanoreview.net/en/cpu-compare/intel-core-i5-1245u-vs-apple-m1} with 8.0 GB of RAM, running Sequoia 15.2. Both maximum clique solvers are run in parallel using 8 threads.

The results are demonstrated in Fig. \ref{fig:time cost}, and the discussion is as follows: (1) For general cases, the average time cost for the entire process with 100 correspondences is approximately 2 ms, demonstrating high efficiency for moderate problem size. Even with 1000 correspondences, the process achieves a time cost of approximately 0.17 s, ensuring it is suitable for real-time implementation. (2) For coplanar cases, the average time cost with 100 correspondences is approximately 5 s, which is somewhat slow for real-time applications. As the number of correspondences increases, the time cost becomes unaffordable, requiring approximately 60 s for a single run with only 150 correspondences. (3) Additionally, we observed that for both cases, the compatibility tests initially account for the majority of the time cost. However, as the number of correspondences increases, the time consumed by the maximum clique solver gradually becomes more significant. For general cases, the time cost is acceptable when the problem size is not excessively large ($\le$ 1000), but for larger sizes, heuristics are needed to accelerate the process. For coplanarity cases, a more efficient $k$-uniform hypergraph maximum clique solver is required, which has not received much research attention to date.

\section{Future Work}

(1) More experiments with diverse settings, including real-world tests, will be conducted to further evaluate the proposed method. (2) A more efficient $k$-uniform hypergraph maximum clique solver will be explored to benefit similar research endeavors.


\renewcommand{\theequation}{A\arabic{equation}} 
\setcounter{figure}{0}
\renewcommand{\thefigure}{A\arabic{figure}}
\setcounter{equation}{0} 
\section*{Appendix A}
Consider two noiseless measurements $\mathbf{m}_i = [r_i, \theta_i]^T$ and $\mathbf{m}_j = [r_j, \theta_j]^T$, and the respective $\phi_i$, $\phi_j$ of their corresponding $\mathbf{p}^s_i$, $\mathbf{p}^s_j$ are confined within $[\phi_{\text{min}},\phi_{\text{max}}]$. The spherical representation is used to align with the sonar noise model, and we directly consider the range of $\| \vv{\mathbf{p}^s_{ij}}\|$, noting that $\| \vv{\mathbf{p}^s_{ij}}\| = \| \vv{\mathbf{p}^w_{ij}}\|$ under a rigid body transformation \cite{shi2021robin, yang2020teaser}. We have
\begin{equation}
	\begin{aligned}[b]
		\| \vv{\mathbf{p}^s_{ij}}\| = \big( (r_i\sin\theta_i\cos\phi_i-r_j\sin\theta_j\cos\phi_j)^2 + \\
		(r_i\cos\theta_i\cos\phi_i - r_j\cos\theta_j\cos\phi_j)^2+ \\
		(r_i\sin\phi_i - r_j\sin\phi_j)^2 \big)^{\frac{1}{2}} \\ 
		\iff \big(r_i^2 + r_j^2 - 2r_i r_j(\cos(\theta_i-\theta_j)\cos\phi_i \cos\phi_j+\\
		 \sin\phi_i \sin\phi_j) \big)^{\frac{1}{2}}.
	\end{aligned}
\end{equation}
Given that the range measurements are determined, we focus on the right multiplier of $-2r_ir_j$. 

\textbf{Maximum Value.} To find the maximum of the right multiplier, we perform a transformation on it and obatin
\begin{multline}
		\overbrace{\cos(\theta_i-\theta_j)\cos\phi_i \cos\phi_j+\sin\phi_i \sin\phi_j}^{A} \iff \\
		\hfil \underbrace{\big(\cos(\theta_i-\theta_j)-1\big)\cos\phi_i \cos\phi_j}_{B}
		 + \underbrace{\cos(\phi_i-\phi_j)}_{C}.
\end{multline}
According to the subadditivity, we have
\begin{equation}
	\max (A) = \max (B+C) \le \max (B) + \max (C).
	\label{equ:subadditivity}
\end{equation}
As mentioned in Sec. \ref{subsec:projection}, it is typically assumed that $\phi_\text{min} = \phi_\text{max} \le 10^\circ$. Then, since $\cos(\theta_i-\theta_j)-1 \le 0$, to maximize $B$, it is necessary to minimize $\cos\phi_i \cos\phi_j$. This can be achieved when $\phi_i, \phi_j \in \{\phi_\text{min},\phi_\text{max}\}$. To maximize $C$, we only need to enforce $\phi_i - \phi_j = 0$. Taking the intersection of the two cases, we get $\phi_i = \phi_j \in \{\phi_\text{min},\phi_\text{max}\}$. Since the intersection is not an empty set, the equality in Eq. \eqref{equ:subadditivity} can be achieved, and the maximum value is $\big(\cos(\theta_i-\theta_j)-1\big)\cos^2\phi_\text{max}+1$. In the degenerate case where $\theta_i = \theta_j$, the maximum value is 1, which occurs when $\phi_i = \phi_j$.

\textbf{Minimum Value.} To find its minimum, we perform another transformation on $A$:
\begin{multline}
	A \iff \overbrace{\cos(\theta_i-\theta_j)\cos(\phi_i-\phi_j)+}^{D} \\ \hfil
	\underbrace{\big(1- \cos(\theta_i-\theta_j)\big)\sin\phi_i \sin\phi_j}_{E}.
	\label{equ:minimum}
\end{multline}
Similarly, we wish $D$ and $E$ can be simultaneously minimized by a non-empty set. If $\cos(\theta_i-\theta_j) \ge 0$, skipping trivial steps, we have $\phi_i = -\phi_j \in \{\phi_\text{min}, \phi_{\text{max}}\}$. If $\cos(\theta_i-\theta_j) < 0$, further discussion is needed.

Without loss of generality, suppose $\phi_j \in [-|\phi_i|,|\phi_i|]$. If not, swapping the $\phi_i$ and $\phi_j$ will again satisfy the condition. Taking the partial derivative of $A$ with respect to $\phi_j$, we get
\begin{multline}
		\frac{\partial A}{\partial \phi_j} = \cos(\theta_i - \theta_j)(\sin\phi_i \cos\phi_j - \sin\phi_j \cos\phi_i) + \\ \hfil
		\big(1-\cos(\theta_i - \theta_j)\big)\sin\phi_i \cos\phi_j.
\end{multline}
If $\phi_i=0$, $\partial A / \partial \phi_j=0$. If $\phi_i>0$, check if $\forall \phi_j \in [-\phi_i,\phi_i], \partial A / \partial \phi_j > 0$, we have
\begin{equation}
	\begin{aligned}[b]
		\cos(\theta_i - \theta_j)(1 - \frac{\tan\phi_j}{\tan\phi_i}) + 
		1-\cos(\theta_i - \theta_j) >0 \\
		\textcolor{gray}{(\phi_i>0 \land \phi_j < 10^\circ)} \\
		\iff 1-\cos(\theta_i - \theta_j)\frac{\tan\phi_j}{\tan\phi_i} > 0 \\
		\iff \tan\phi_j > \frac{\tan\phi_i}{\cos(\theta_i - \theta_j)} \\
		\textcolor{gray}{(\cos(\theta_i - \theta_j) < 0)}.
	\end{aligned}
\end{equation}
The inequality always holds, as 
\begin{equation}
	\tan\phi_j \ge \tan(-\phi_i) > \frac{\tan\phi_i}{\cos(\theta_i - \theta_j)}.
\end{equation}
Therefore, in this case, when $A$ reaches its minimum, $\phi_j=-\phi_i$. The conclusion remains the same for $\phi_i < 0$ due to the symmetry characteristics inherent in 2D FLS imaging. Take them back into Eq. \eqref{equ:minimum}, we have
\begin{equation}
	\cos(\theta_i-\theta_j) - \big(1+\cos(\theta_i-\theta_j)\big)\sin^2{\phi_i}.
\end{equation}
Since $- \big(1+\cos(\theta_i-\theta_j)\big) \le 0$, the minimum is reached when $\phi_i = -\phi_j \in \{\phi_\text{min}, \phi_{\text{max}}\}$, which aligns with the other cases. At this point, the minimum value is $\cos(\theta_i-\theta_j) - \big(1+\cos(\theta_i-\theta_j)\big)\sin^2{\phi_\text{max}}$.

\textbf{Summary.} As the range measurements are always greater than $0$, it follows that $-2r_ir_j < 0$. Consequently, the maximum and minimum values of $\| \vv{\mathbf{p}^s_{ij}} \|$ are achieved under opposite conditions.

\section*{Appendix B}
\label{app:noisy length constraint}
\textbf{Lower Bound under Noisy Measurements.} Considering noisy measurements in the lower bound of Eq. \eqref{equ:length bounds}, we have
\begin{multline}
	\bigg( (r_i + \eta_{i})^2 + (r_j + \eta_{j})^2 - 2(r_i + \eta_{i})(r_j + \eta_{j})\cdot
	\\
	\bigg(\big(\cos(\theta_i-\theta_j + \epsilon_i - \epsilon_i)-1\big)\cos^2\phi_\text{max}+1\bigg)\bigg)^{\frac{1}{2}}.
	\label{equ:lower bound noisy}
\end{multline}
Since $\eta$ is independent of $\epsilon$, and assuming $- 2(r_i + \eta_{i})(r_j + \eta_{j}) < 0$ (as the sonar does not return non-positive distance measurments), we can initially focus on determining the maximum of the term involving only $\theta$, i.e. the right multiplier of the expression $- 2(r_i + \eta_{i})(r_j + \eta_{j})$. We adopt a bounded noise assumption, expressed as $|\epsilon| \le \beta_\theta$, to facilitate the derivation. The value of $\beta_\theta$ can be set to $3\sigma_\theta$ or a more conservative estimate. It should be noted that $\beta_\theta \ll \theta_\text{max} - \theta_\text{min}$ always holds true as specified in the sonar's manual. Under this assumption, we have $|\epsilon_i - \epsilon_j| \le 2\beta_\theta$. Considering the typical range of \(\theta\) (Sec. \ref{subsec:projection}) and applying the triangle inequality, we have:
\begin{multline}
	|\theta_i-\theta_j + \eta_{\theta_i} - \eta_{\theta_i}| \le |\theta_i-\theta_j| + |\eta_{\theta_i} - \eta_{\theta_i}| \le \\ |\theta_i-\theta_j| + 2\beta_\theta < 180^\circ.
\end{multline}
Then, we only need to select the value closest to $0$ to maximize $\cos(\cdot)$. The result of the right multiplier is denoted by $\mathbf{X}_{\text{min}}^*$. It is straightforward to verify that $-1 < \mathbf{X}_{\text{min}}^*\leq 1$, which can be interpreted as the value of $\cos\alpha$. Here, $\alpha$ is the \textbf{virtual} angle between two vectors with magnitude of $r_i$ and $r_j$ in the 2D plane. A triangle is formed in this interpretation. Then, Eq. \eqref{equ:lower bound noisy} represents the length of the side opposite to $\alpha$. Since $\alpha$ is determined, and the lengths of the corresponding adjacent sides are influenced by noise, the feasible regions for the far endpoints become two line segments, each with a length of $2\beta_r$, where $|\eta| \leq \beta_r$ under the bounded noise assumption. The problem then converts to finding the smallest distance between these two segments, which has been well discussed in \cite{lumelsky1985fast}.

\textbf{Upper Bound under Noisy Measurements.} For the upper bound, the route of analysis is the same as for the lower bound condition, except that minimizing the length of the opposite side becomes a maximization problem. The maximum length can always be achieved at the endpoints of the feasible regions of the two adjacent sides. The proof is trivial, so we omit it for conciseness.

\begin{figure}[h]
	\centering
	\begin{tikzpicture}
		\node[anchor=south west,inner sep=0] (image) at (0,0) 
		{\includegraphics[width=\linewidth]{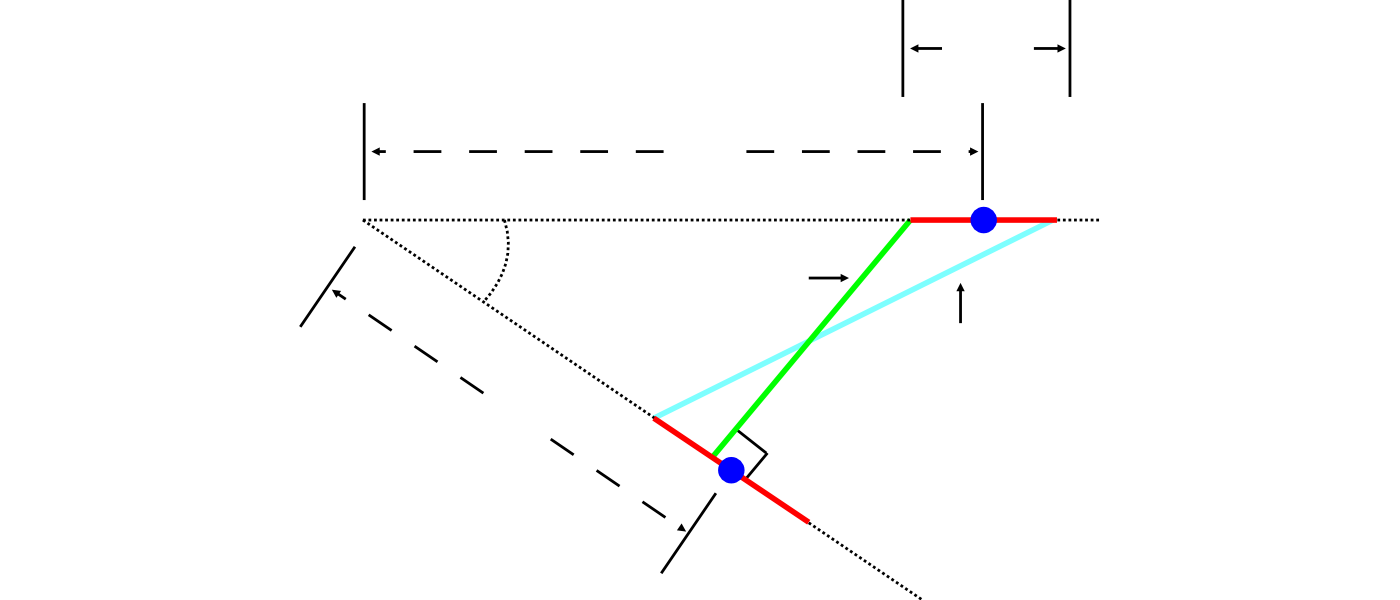}};
		\begin{scope}[shift={(image.south west)}, x={(image.south east)}, y={(image.north west)}]
			\node at (0.69,0.4) {max};
			\node at (0.54,0.55) {min};
			\node at (0.51,0.74) {$r_i$};
			\node at (0.37,0.3) {$r_j$};
			\node at (0.71,0.92) {2$\beta_{r}$};
			\node at (0.39,0.55) {$\alpha$};
		\end{scope}
	\end{tikzpicture}
	\caption{The problem of determining the lower and upper bounds under noisy measurements is transformed into searching for the minimum and maximum lengths between two segments, which are defined by $\beta_r$, $r_{\{i,j\}}$, and $\alpha$ (determined from $\theta_{\{i,j\}}$).}
	\label{fig:length constraint noisy illustration}
\end{figure}

Fig. \ref{fig:length constraint noisy illustration} provides an intuitive explanation of the geometric meaning for the aforementioned derivation.

\section*{Appendix C}
By explicitly expressing $r$ and $\theta$ in Eq. \eqref{equ:coplanarity check} and setting $b_4 = -1$, for the first row, we obtain
\begin{equation}
	\label{equ:coplanarity check first row}
	\begin{aligned}[b]
		 &\sum_{i=1}^4b_i(r_i+\eta_i)\big(\sin(\theta_i+\epsilon_i)\big)
		\\
		&\textcolor{gray}{\text{(First-order Taylor expansion.)}}
		\\
		\approx
		&\sum_{i=1}^4b_i
		(r_i+\eta_i)(\sin\theta_i+\epsilon_i \cos\theta_i) \iff
		\\
		&\sum_{i=1}^4b_i(r_i\sin\theta_i+\epsilon_i r_i\cos\theta_i+\eta_i\sin\theta_i+\eta_i\epsilon_i\cos\theta)
		\\
		&\textcolor{gray}{(\eta \cdot \epsilon \approx 0)}
		\\
		\approx
		&\sum_{i=1}^4b_i(r_i\sin\theta_i+\epsilon_i r_i\cos\theta_i+\eta_i\sin\theta_i).
	\end{aligned}
\end{equation}
Through the additivity of the Gaussian distribution, the result of Eq. \eqref{equ:coplanarity check first row} follows
\begin{equation}
	\mathcal{N}\left (\sum_{i=1}^4 b_ir_i\sin\theta_i, \sum_{i=1}^4b_i^2(r_i^2\cos^2\theta \sigma_\theta^2+\sin^2\theta\sigma_r^2)\right ).
\end{equation}
The mean is exactly the original noiseless residual. If all four correspondences are correct, the residual is zero, yielding a zero-mean Gaussian. The second row's derivation is similar, with $\sin$ replaced by $\cos$.


\bibliographystyle{IEEEtran}
\bibliography{ref}

\addtolength{\textheight}{-12cm}   
\end{document}